\documentclass[10pt,conference,final]{IEEEtran}

\usepackage{amsmath}
\usepackage{enumerate}
\usepackage{graphicx}
\usepackage[utf8]{inputenc}
\usepackage{subfigure}
\usepackage{varioref}
\usepackage{xcolor}
\usepackage{booktabs}
\usepackage{tikz}

\newcommand{\ft}[1]{\footnotesize{#1 }}

\definecolor{red1}{rgb}{1.0, 0.5, 0.5}
\definecolor{red2}{rgb}{1.0, 0.7, 0.7}
\definecolor{red3}{rgb}{1.0, 0.9, 0.9}

\definecolor{black}{rgb}{0.0, 0.0, 0.0}
\definecolor{grey1}{rgb}{0.7, 0.7, 0.7}
\definecolor{grey2}{rgb}{0.9, 0.9, 0.9}

\definecolor{purple1}{rgb}{0.6, 0.0, 0.8}
\definecolor{purple2}{rgb}{0.91, 0.5, 1.0}
\definecolor{purple3}{rgb}{0.98, 0.9, 1.0}

\definecolor{blue1}{rgb}{0.0, 0.35, 0.7}
\definecolor{blue2}{rgb}{0.6, 0.79, 1.0}
\definecolor{blue3}{rgb}{0.9, 0.95, 1.0}

\newcommand{\best}[1]{\colorbox{blue1}{\color{white} #1}}
\newcommand{\five}[1]{\colorbox{blue2}{#1}}
\newcommand{\ten}[1]{\colorbox{blue3}{#1}}

\newcommand{\sub}[1]{$_{\text{#1}}$}

\begin{document}

\title{Monte Carlo Methods for the Game Kingdomino}

\author{\IEEEauthorblockN{Magnus Gedda, Mikael Z. Lagerkvist, and Martin Butler}
\IEEEauthorblockA{Tomologic AB\\
Stockholm, Sweden\\
Email: firstname.lastname@tomologic.com}
}

\maketitle

\begin{abstract}

\emph{Kingdomino} is introduced as an interesting game for studying
game playing: the game is multiplayer (4 independent players per
game); it has a limited game depth (13 moves per player); and it has
limited but not insignificant interaction among players.

Several strategies based on locally greedy players, Monte Carlo
Evaluation (MCE), and Monte Carlo Tree Search (MCTS) are presented
with variants. We examine a variation of UCT called
\emph{progressive win bias} and a playout policy
(\emph{Player-greedy}) focused on selecting good
moves for the player. A thorough evaluation is done showing how the
strategies perform and how to choose parameters given specific time
constraints. The evaluation shows that surprisingly MCE is stronger
than MCTS for a game like Kingdomino.

All experiments use a cloud-native design, with a game server in a
Docker container, and agents communicating using a REST-style JSON
protocol. This enables a multi-language approach to separating the
game state, the strategy implementations, and the coordination layer.

\end{abstract}

\begin{IEEEkeywords}
Artificial intelligence, games, Monte Carlo, probabilistic computation,
heuristics design.
\end{IEEEkeywords}

\section{Introduction}

Implementations and heuristics for computer players in classical board
games such as Chess, Go and Othello have been studied extensively in
various contexts. These types of games are typically two-player,
deterministic, zero sum, perfect information games.  Historically,
game theoretic approaches such as Minimax and similar variants such as
Alpha-Beta pruning have been used for these kinds of games, dating back
to Shannon in 1950~\cite{Shannon1950}.  Recently more advanced
techniques utilizing Monte Carlo methods~\cite{Metropolis1949} have
become popular, many of them outperforming the classical game
theoretic approaches~\cite{Abramson1990, Bouzy2004, Silver2016}.

The characteristics of the Monte Carlo-based methods also make them
viable candidates for games with more complex characteristics such as
multiplayer, nondeterministic elements, and hidden
information~\cite{Sturtevant2008}. With the recent emergence of more
modern board games (also called \emph{eurogames}), which often exhibit
these characteristics, we naturally see more and more research
successfully applying Monte Carlo-based methods to such
games~\cite{Szita2009, Nijssen2012, Robilliard2014, Browne2012}.

Among the most common Monte Carlo-based methods we have Monte Carlo
Evaluation (MCE) (also called \emph{flat
  Monte Carlo})~\cite{Abramson1990} and Monte Carlo Tree Search
(MCTS)~\cite{Kocsis2006, Coulom2006}. Flat Monte Carlo has shown some
success~\cite{Bouzy2004} but is generally considered too slow for games
with deep game trees~\cite{Nijssen2013}. MCTS has come to address the
problems of MCE and become a popular strategy for modern board
games. A plethora of enhancements have been presented for MCTS, both
general and domain-dependent, increasing its performance even further
for various games~\cite{Schaeffer1983, Winands2004, Gelly2007,
  Chaslot2008, Nijssen2010}. For shallow game trees it is still
unclear which Monte Carlo method performs best since available
recommendations only concern games with deep trees.

Kingdomino~\cite{game:kingdomino} is a new board game which won the
prestigious \emph{Spiel des Jahres} award 2017. Like many other
eurogames it has a high branching factor but differs from the general
eurogame with its shallow game tree (only 13 rounds). It has frequent
elements of nondeterminism and differs from zero sum games in that the
choices a player makes generally have limited effect on its
opponents. The game state of each round can be quantified to get a
good assessment of how well each player is doing which facilitates
strong static evaluators. The difference in characteristics compared
to previously examined eurogames can potentially render previous
recommendations misleading.

We examine static evaluators, Monte Carlo Evaluation (MCE) and
Monte Carlo Tree Search using the Upper Confidence Bound for Trees
algorithm (UCT). Vanilla implementations of MCE and UCT are compared
with various enhancements such as heuristics for more realistic
playout simulations and an improvement to UCT which initially steers
the selection towards more promising moves. All variants are
thoroughly evaluated showing how to select good parameters.

The experimental focus is on heuristic design rather than building
efficient competitive agents, i.e., the implementations are meant to
be comparative rather than relying on low-level optimization tweaks.
All agents are independent processes communicating with a game server
using a JSON protocol.

\section{Kingdomino}
\label{sec:kingdomino}

Kingdomino~\cite{game:kingdomino} is a modern board game for 2-4
players released in 2016 where the aim of each player is to expand a
kingdom by consecutively placing \emph{dominoes} provided in a
semi-stochastic manner. A domino contains two \emph{tiles}, each
representing a \emph{terrain} type and can have up to three
\emph{crowns} contributing to the score for its area. The goal is to
place the dominoes in a 5x5 grid with large areas connecting terrains
of the same type (using 4-connectivity) containing many crowns to
score points.

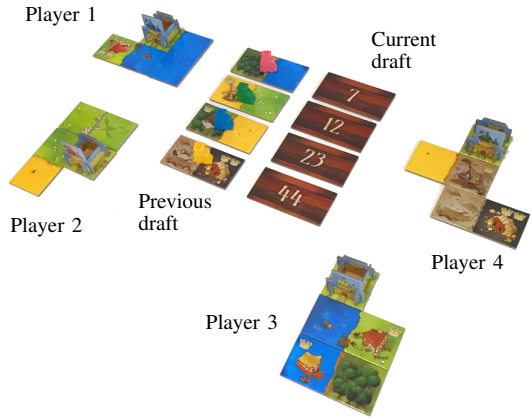
\begin{figure}[!t]
\centering
\begin{tikzpicture}
  \draw (0, 0) node[inner sep=0] {\includegraphics[width=8cm]{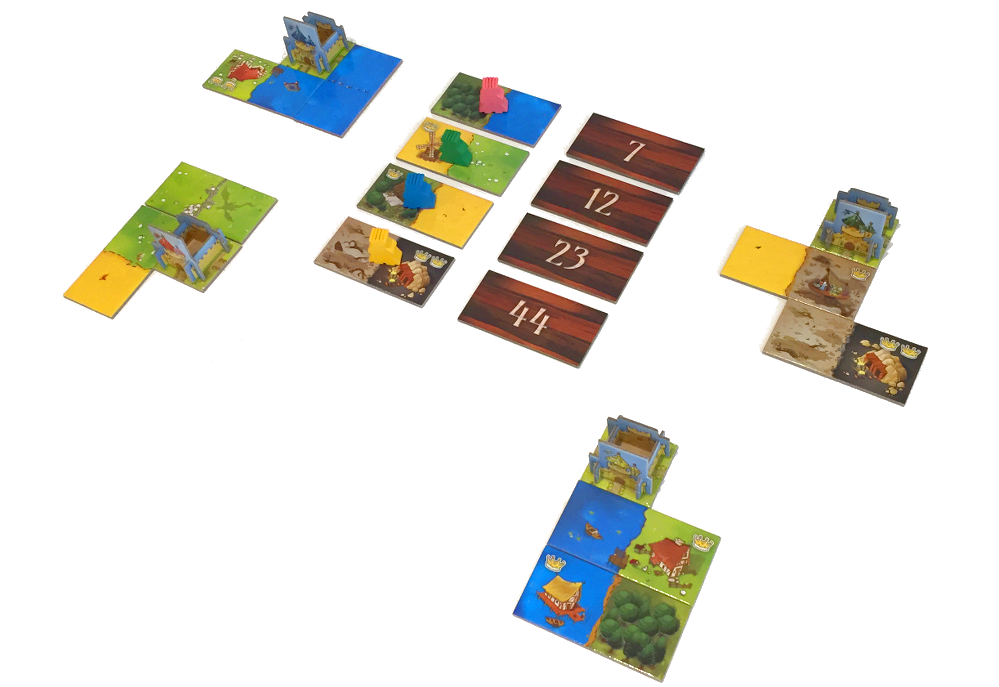}};
  \draw (-2.8, 2.6) node {\ft{Player 1}};
  \draw (-3.0, -0.2) node {\ft{Player 2}};
  \draw (-0.4, -1.5) node {\ft{Player 3}};
  \draw (2.6, -0.7) node {\ft{Player 4}};
  \draw (-1.9, 0.15) node [anchor=west]{\ft{Previous}};
  \draw (-1.9, -0.15) node [anchor=west] {\ft{draft}};
  \draw (1.2, 2.3) node [anchor=west]{\ft{Current}};
  \draw (1.2, 2.0) node [anchor=west] {\ft{draft}};
\end{tikzpicture}
\caption{Kingdomino in-game setup}
\label{fig:kingdomino_setup}
\end{figure}

\subsection{Rules (3-4 Players)}
\label{sec:kingdomino:rules}

You begin with your castle tile placed as the starting point of your
kingdom and a meeple representing your king.  In the first round, the
same number of dominoes as there are kings in play are drawn from the
\emph{draw pile} and added to the \emph{current draft}. Each player
then chooses one domino each from the current draft by placing their
king on the chosen domino. When all dominoes in the draft have been
chosen, the game moves on to the second round by drawing a new current
draft from the draw pile. The previous current draft (the one that now
has a king on each domino) becomes the \emph{previous draft}.

In round two, and every consecutive round up until the last, the
player with the king placed on the first domino in the previous draft
adds the chosen domino to their territory, according to the connection
rules, and chooses a new domino from the current draft by placing the
king on the chosen domino. The other players then do the same
placement-selection move in the order their kings are positioned in
the previous draft.  A placed domino must either connect to the castle
tile or another domino matching at least one of its terrains
(horizontally or vertically only). If you cannot add a domino to your
kingdom, the domino will be discarded.

The last round works the same as the previous rounds with the
exception that there are no more dominoes to draw from the draw pile
and therefore there will be no current draft from which to choose any
new dominoes.

The final score is the sum of the scores for each 4-connected area of
the same terrain type. The score for each area is the number of tiles
multiplied by the total number of crowns on the area. Note that for an
area with no crowns, the score is zero.  There are also two additional
rules used in this paper (both part of the official game rules). The
first is the \emph{Middle Kingdom} rule, which states that you get an
additional 10 points if your castle is in the center of the 5x5
grid. The second is the \emph{Harmony} rule, which states that you get
an additional 5 points if your territory is complete (i.e., no
discarded dominoes).

For a complete description of the rules, including rules for 2
players, we refer to~\cite{game:kingdomino}.

\subsection{Game characteristics}
\label{sec:kingdomino:characteristics}

Kingdomino is classified as a non-deterministic game since the
dominoes are drawn randomly from the draw pile. All players have a
similar goal and all players have complete information of the game
state at all times, which means that it is also a symmetric perfect
information game.

The number of possible draws from the deck is defined by the following formula.
$
\prod_{i=0}^{11} {48 - 4i \choose 4}
\approx
3.4\cdot10^{44}
$
The most interesting thing about the number of possible draws is that
it is significantly less than the total number of shuffles of the deck
(around a factor of $3.6\cdot10^{16}$).

\begin{figure}[!t]
\centering
\includegraphics{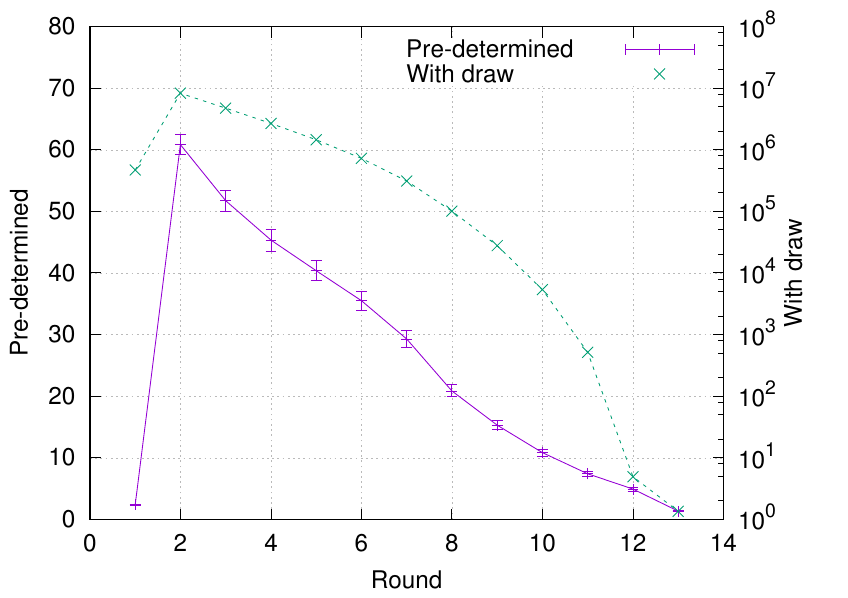}
\caption{Average branching factor per round for a random player when
  playing against three random opponents ($1000$ games). The error
  bars show the 95\% confidence interval.}
\label{fig:experiment1_branching_factors}
\end{figure}

\figurename~\ref{fig:experiment1_branching_factors} shows the
branching factor for each round. This is computed experimentally using
4-player games with the players choosing moves randomly (see
Section~\ref{sec:experiments:setup}). Assuming that the branching
factor for player $p$ in round $r$ is an independent stochastic
variable $B_{pr}$, multiplying the expected value for the branching
factor each round gives the expected value for the game tree size
\emph{given a predetermined deck shuffle}.  Using the experimentally
determined values for $B_{pr}$, the game tree size is approximately
$$
E\left[\prod_{p=1}^4\prod_{r=1}^{13} B_{pr}\right] = \prod_{p=1}^4\prod_{r=1}^{13} E[B_{pr}] 
\approx 3.74\cdot10^{61}
$$

When accounting for the number of possible draws from the deck, the
number of Kingdomino games is around $1.27\cdot 10^{106}$. This puts
Kingdomino at a game tree complexity between Hex and Chess when
accounting for all shuffles, and similar to Reversi/Othello for a
pre-determined shuffle~\cite{wikipedia:game-tree-complexity}.

\section{Strategies for Kingdomino}
\label{sec:strategies}

Agents can be implemented using a wide range of strategies. Here we
focus on statistical evaluators such as Monte Carlo Evaluation and
Monte Carlo Tree Search together with various enhancements. We also
include some static evaluators to analyse game characteristics and use
as reference agents when evaluating the statistical strategies.

\subsection{Static Evaluators}

Kingdomino generally has a solid score progression which makes it
feasible to implement strong static evaluators by computing the score
of each player at every state of the game, unlike, e.g., the game of
Go which has to rely heavily on statistical methods since
domain-dependent move generators are very difficult to
improve~\cite{Bouzy2004}. Also, considering Kingdomino is a perfect
information game, any static evaluator with a greedy approach could
potentially be competitive. We define two static evaluators,
\emph{Greedy Placement Random Draft} (GPRD) and \emph{Full Greedy}
(FG). GPRD places each domino in a greedy manner (to get maximum point
increase) but selects dominoes randomly from the current draft while
FG uses both greedy placement and selects greedily from the current
draft. Both evaluators avoid moves that break the \emph{Middle
  Kingdom} rule or result in single-tile holes. The FG evaluator is
likely to act similar to an above average human player since it
incorporates the visible domain knowledge to make realistic moves
without using any search strategies.

\subsection{Monte Carlo Methods}

Monte Carlo methods such as Monte Carlo Evaluation
(MCE)~\cite{Abramson1990} and Monte Carlo Tree Search
(MCTS)~\cite{Kocsis2006, Coulom2006} have recently been used
successfully for building computer players in both classical
two-player deterministic board games, such as Go~\cite{Bouzy2004}, and
more modern multiplayer non-deterministic board games, such as
Settlers of Catan~\cite{Szita2009}, Scotland Yard~\cite{Nijssen2012},
and 7 Wonders~\cite{Robilliard2014}.

\subsubsection{Monte Carlo Evaluation}

In flat Monte Carlo search (which we in this paper refer to as
\emph{Monte Carlo Evaluation}), each game state is represented by a
node in a tree structure and the edges represent possible moves. The
root node represents the current game state and its children represent
the game states produced by each available move. The evaluation
selects a child node randomly (using uniform sampling) and simulates a
complete game from that node (referred to as a \emph{playout}), using
some \emph{playout policy}, until termination. The selection-playout
procedure is done repeatedly until either a maximum number of playouts
have been reached or the time runs out. Each child node stores the
average result from all its playouts, and the the max child is
selected as the best move. Evaluators based on MCE have shown to be
strong players in small classical games, such as 3x3 Tic-Tac-Toe, and
play on par with standard evaluators on larger
games~\cite{Abramson1990}.

The high exponential cost of searching trees with high branching
factors makes global tree search impossible, especially under tight
time constraints. However, the search depth of Kingdomino is shallow
enough for MCE to potentially be a viable option since a shallow game
tree facilitates high termination frequencies even at early stages in
the game.

\subsubsection{Monte Carlo Tree Search}

Monte Carlo Tree Search expands on the functionality of Monte Carlo
Evaluation by expanding the search tree asymmetrically in a best-first
manner guided by statistics. A commonly used Monte Carlo Tree search
algorithm for game play is UCT~\cite{Kocsis2006}, which guides the
search by computing the Upper Confidence Bound (UCB) for each node and
select moves for which the UCB is maximal. The UCB is defined as
\begin{equation}
\label{eqn:ucb}
\mathrm{UCB} = \bar{X}_i + C\sqrt{\frac{\ln T}{T_i}},
\end{equation}
where $\bar{X}_i$ is the average payoff of move $i$, $T$ is the number
of times the parent of $i$ has been visited, $T_i$ is the number of
times $i$ has been sampled, and $C$ is the exploration constant. For a
full description of the UCT algorithm we refer
to~\cite{Kocsis2006}. UCT, with enhancements such as domain-specific
heuristics in the playout policies, has been shown to perform well for
games with high branching factors~\cite{Sturtevant2008}.

\subsection{Playout Policy Enhancements}

The playout policy in its standard form uses random move selection
throughout the playout. A common enhancement is to incorporate,
potentially time expensive, domain-dependent heuristics to get more
realistic playouts. We examine four different playout policies. The
true random playout policy (TR) which chooses all moves randomly in
the playout. The $\epsilon$-greedy policy
($\epsilon$G)~\cite{Sturtevant2008} which chooses moves randomly with
$\epsilon$ probability and greedily with probability
$(1-\epsilon)$. The full greedy policy (FG) which chooses all moves
greedily. And finally we use a playout policy we call the
\emph{player-greedy} policy (PG). It chooses the player's move
greedily and all opponent moves randomly. Random opponent modelling
has recently been applied successfully in multi-player tracks of
General Video Game Playing (GVGP) AI competitions~\cite{Gaina2017}
but has, to our knowledge, not previously been applied to AI in board
games. The player-greedy policy should be favourable in Kingdomino
since the actions of the opponents generally have limited (but not
insignificant) impact on the player. Its success in the GVGP setting
can likely be attributed to the tight time constraints for opponent
modelling in GVGP.

The $\epsilon$-greedy and player-greedy strategies combine the
advantage of domain knowledge with the speed provided by random move
selection. With a low branching factor, there is a reasonable chance
that good moves will be made with some frequency in random
sampling. But games with large branching factors, such as Kingdomino,
generally have many irrelevant, or even detrimental, moves. In
these games the probability of playing out good moves during random
playouts is relatively small, so there should be a large benefit to
using informed simulation strategies.

\subsection{Scoring Functions}

The \emph{scoring function} defines how the result of a playout is
measured. The basic scoring function is the \emph{Win Draw Loss}
function (WDL) which simply gives a winning playout the score $1$, a
playout where the player is tied with an opponent for first place (a
draw) the score $0.5$, and a playout which is not a win or a draw the
score $0$. The reward model in Monte Carlo Evaluation facilitates more
sophisticated scoring functions. One such function, which we refer to
as the \emph{Relative} scoring function (R), takes the player's score
relative to the score of the highest scoring opponent $f = p_s/(p_s +
q_s)$, where $p_s$ is the player score and $q_s$ is the opponent
score. A third third scoring function, which we refer to as the
\emph{Player} scoring function (P), simply uses the player's
score. This function does not care whether the player wins or loses
and only tries to maximize the player's own score.

\subsection{MCTS Selection Enhancements}

Among the popular enhancements for MCTS there are learning
enhancements such as RAVE~\cite{Gelly2007} and the history
heuristic~\cite{Schaeffer1983, Winands2004}. They use offline
information from previous games to guide the selection toward moves
that have been successful in past games. Kingdomino has a low
\emph{n}-ply variance which means it could potentially benefit from
learning enhancements~\cite{Sturtevant2008}. However, in Kingdomino
the reward of a single move is dependent on the game state, so the
game state has to be incorporated in the offline information for each
move. This has the effect of drastically decreasing the hit
probability of a move while increasing lookup time.

A popular online enhancement is progressive bias~\cite{Chaslot2008}
which guides the selection towards promising moves by using a --
potentially time consuming -- heuristic value which diminishes with
increasing visits to the node. Here we use a selection enhancement
which we call \emph{progressive win bias} which combines progressive
bias with a tweak that makes the heuristic value diminish with the
number of node losses instead of the number of node visits. The tweak
has successively been applied to the game Lines of
Action~\cite{Winands2010} but has never been evaluated in a systematic
fashion as presented here. We define progressive win bias as
\[
W\frac{H_i}{T_i\left(1 - \bar{X}_i\right) + 1},
\]
where $H_i$ is the heuristic value, $\bar{X}_i$ is the average reward
for the node, $T_i$ is the number of node visits, and $W$ is a
positive constant which controls the impact of the bias. In this paper
we use $H_i = S_i - S_{i-1}$ as heuristic, where $S_{\gamma}$ is the
player's score after move $\gamma$. The formula is simply added to the
regular UCB in~\ref{eqn:ucb}.

\section{Implementation}
\label{sec:implementation}

The implementation for the game is based on a server-client
architecture. The server maintains all current, future, and past
games, while a client agent can play in one or more games. A game is
initiated with a set number of players, putting it in the list of
future games. An agent can join a game, on which it receives a secret
token enabling it to make moves for a player in the game. After enough
players join the game, it is started. The game server has a
graphical front-end showing all current and past games with full
history for analysis and inspection.

Agents poll the server for the current game state: the kingdoms and
their scores; the current and next draft; the current player; all
possible moves; and all previously used dominoes.
To make a move, the agent for the current player chooses one of the
possible moves. The communication is based on a HTTP REST JSON
API. The protocol gives enough information to enable stateless
agents that only need remember their secret token. When joining a
game, it is possible for an agent to register an HTTP callback
endpoint that the server uses to notify the agent that its player is
the current player.

The game server is implemented in Scala, and is packaged as a Docker
container. This simplifies running the server in any setting, either
on a remote server or locally. In particular, the choice of using
standard web technologies for communication leads to a clean and
simple separation of agents and the server.

At a one-day hackathon, 7 programmers could without preparation build
rudimentary game playing agents in a variety of languages (Java,
Scala, Python, Rust, and Haskell). The state representation and the
full valid move list make it simple to implement static evaluators,
without having to implement the full game logic. Naturally, for a more
competitive client the full game logic needs to be implemented also in
the client.

\section{Experiments}
\label{sec:experiments}

Our experiments are intended to give insights into the game, to give
guidance on what strategies and algorithms are useful, and how to tune
parameters for the strategies. To compare strategies, we have made the
choice to use static time limits per ply to study how well different
strategies can make use of a specific time allotment without
introducing the complexities of full time management.

Note that all games in these experiments are 4-player games (unless
otherwise stated), so a when a strategy plays equally well as its
opponent it will result in a 25\% win rate. All intervals (in both
figures and tables) represent the 95\% confidence interval.

In board games the number of victories alone can be considered
insufficient to determine the strength of a player. This is supported
by the USOA (United States Othello Association) which uses the margin
of victory as the single most important feature in determining a
player's rating~\cite{Abramson1990}. Therefore, most of our
experiments use the victory margin to determine player strength.

\subsection{Setup}
\label{sec:experiments:setup}

All agents used in the experiments are written in Java and run on a
single threaded 3.2 GHz Intel Core i7 with 12 GB RAM that is also
running the game server. While the agents are not written to be the
fastest possible, some care has been taken to keep the implementation
reasonably fast. The goal is to facilitate comparison between the
agents, not to implement a certain algorithm optimally.

\subsection{Agents}
\label{sec:experiments:agents}

We use three different static evaluator agents: the True Random (TR)
agent, the Greedy Placement Random Draft (GPRD) agent, and the Full
Greedy (FG) agent. The FG agent is used as reference player against
which we evaluate all statistical players.

Each Monte Carlo Evaluation agent is implemented using flat Monte
Carlo search and characterized by a playout policy/scoring function
combination. We denote them by MCE-$X$/$Y$ where $X$ is the playout
policy and $Y$ is the scoring function.

The Monte Carlo Tree Search agents all use the WDL scoring function
and are therefore only characterized by playout policy and selection
enhancements. The MCTS agents lack the possibility of using a relative
scoring function but use maximum score increase as tie breaker
for moves of equal win rate. We denote the standard MCTS agents by
UCT-$X$, the MCTS agents using progressive bias by UCT\sub{B}-$X$, and
progressive win bias by UCT\sub{W}-$X$, where $X$ is the playout
policy.

\subsection{Impact of Domain Knowledge}

In the first experiment we wanted to quantify how basic domain
knowledge affects strategies based on static evaluators. We did this
by playing a True Random player (TR), a Greedy Placement Random Draft
player (GPRD), and a Full Greedy player (FG) 1000 games each against
three TR opponents and registered the number of wins, draws, and
losses. We also registered the score after each round in every game to
see the general score progression of each strategy.

\begin{figure}[!t]
\centering
\includegraphics{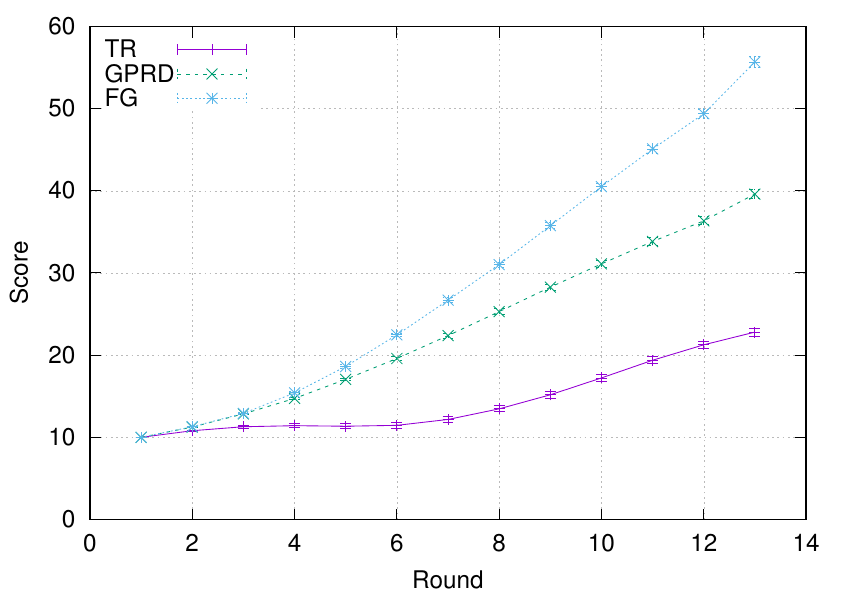}
\caption{Average scores against three TR opponents (1000 games).}
\label{fig:experiment1_scores}
\end{figure}

The average score progression for the three different strategies over
is shown in \figurename~\ref{fig:experiment1_scores}.
All players start with 10p since the castle is within three tiles
distance from the tile furthest away, thus fulfilling the Middle
Kingdom rule. We can clearly see that the TR player had trouble
increasing its score and even dipped around Round 5-6 due to breaking
the Middle Kingdom rule. The GPRD player did a better job, showing
that it is of great importance to select good positions for the placed
domino. However, the score progression of the FG player indicates that
it is of equal importance to also select a good domino from the
current draft (the score for FG is approximately twice the score of
GPRD when corrected for the scores of random moves).

The number of wins, losses, and draws for each strategy are shown in
Table~\ref{tab:experiment1_table1}. Here we see that the FG player
truly outplayed the TR opponents, which was anticipated. More
interesting is that the GPRD player only has approximately 79\% win
rate against the TR opponents. So while carefully selecting
placements, making an uninformed selection from the current draft has
a noticeable impact when played against random opponents.

\begin{figure}[!t]
\centering
\includegraphics{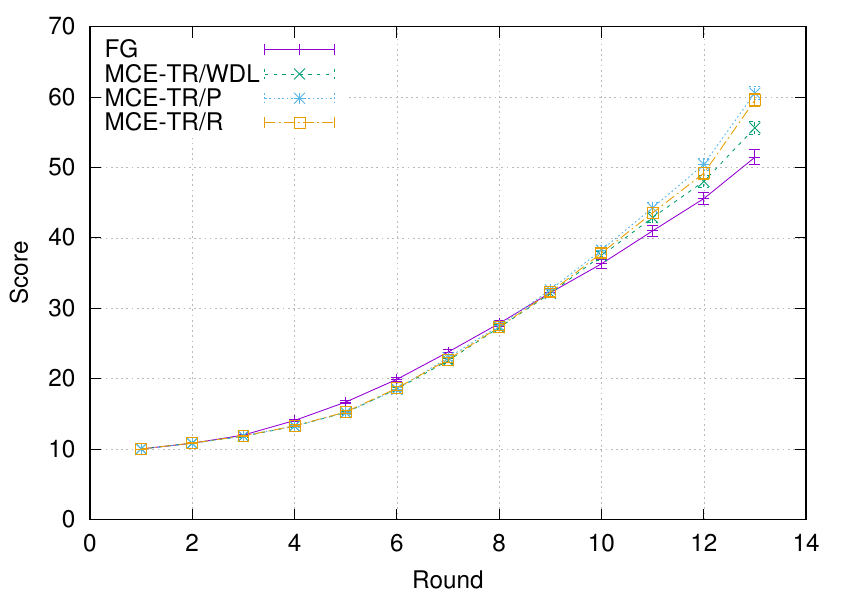}
\caption{Average scores against three FG opponents ($500$ games).}
\label{fig:experiment2_scores}
\end{figure}

\begin{table}[!t]
\begin{center}
\caption{Win percentages for $1000$ games against three TR opponents.}
\label{tab:experiment1_table1}
\footnotesize{
\begin{tabular}{ll*{3}{r}}
\toprule
Player Strategy  & $\,$ & \multicolumn{3}{c}{Opponent Strategy} \\
        & & \multicolumn{3}{c}{TR}   \\
\cmidrule{3-5}
       &  & Wins (\%)  & Draws (\%) & Losses (\%)  \\
\midrule
TR            & & 223 (22.3) & 29 (2.9) & 748 (74.8)  \\
GPRD          & & 794 (79.4) & 22 (2.2) & 184 (18.4)  \\
FG            & & 977 (97.7) & 2 (0.2) & 21 (2.1)  \\
\bottomrule
\end{tabular}
}
\end{center}
\end{table}

\subsection{Static vs Statistical Evaluators}
\label{sec:experiments:searchvsgreedy}

In this experiment we investigated how simple statistical evaluation
performs compared to the best static evaluator-based strategy.  We
also look at how different scoring functions affect the performance of
the statistical evaluators.  We did this by playing three Monte Carlo
Evaluation players, each using a different scoring function and random
selection playout policy, 500 games each against three FG opponents and compared the
results to the same number of games played by a FG player against
three FG opponents. The time limit was set to 5s per ply. The three
Monte Carlo players were MCE-TR/WDL, which only counts the number of
wins/draws/losses and chooses the move that maximises the number of
wins, MCE-TR/P, which tries to maximise the player's final score, and
MCE-TR/R, which tries to maximise the victory margin. The score
progressions are shown in \figurename~\ref{fig:experiment2_scores} and
the final scores in Table~\ref{tab:experiment3_table}.

\begin{table}[!t]
\begin{center}
\caption{Average scores for $500$ games against three FG opponents.}
\label{tab:experiment3_table}
\footnotesize{
\begin{tabular}{ll}
\toprule
Player Strategy & Avg. Score \\
\midrule
FG          & 51.4 (2.1) \\
MCE-TR/WDL  & 55.6 (1.8) \\
MCE-TR/P    & 60.6 (1.9) \\
MCE-TR/R    & 59.5 (1.8) \\
\bottomrule
\end{tabular}
}
\end{center}
\end{table}

The experiment clearly shows that the statistical evaluators
significantly outperform the FG player. It is interesting to see how
the statistical evaluators select sub-greedy moves in the middle of
the game to enable higher payoffs in the later parts of the game. It
is also clear that MCE-TR/WDL does not reach as high final score as
the other statistical evaluators. This is most likely a result of the
WDL scoring function's lack of score information which renders it
incapable of discriminating between branches where all leaf nodes
result in a win while it is in the lead. Since each node only stores
the winning average, it will not be able to determine which branch
will lead to a higher final score. Also, the R and P scoring functions
are more robust against the recurring stochastic events. There is no
significant difference in performance between the Player scoring
function and Relative scoring function.

\subsection{Enhanced Playout Policies}

In this experiment we investigated the effect of different enhancements
to Monte Carlo Evaluation by incorporating domain knowledge into the
playout policies. We did this by playing Monte Carlo Evaluation
players, both with and without domain knowledge, against three FG
opponents and compared the results. The players we used were MCE-TR/R,
which has no domain knowledge at all and only selects moves randomly
for both the player and opponents in the playouts, MCE-$\epsilon$G/R
with $\epsilon=0.75$, which uses random selection in 75\% of the times
in the playout and greedy selection 25\% of the times, MCE-PG/R, which
uses greedy selection for the player and random selection for the
opponents in the playouts, and MCE-FG/R, which uses greedy selection
for all moves in the playouts. We used the relative scoring function
since its goal aligns with the measure of player strength and
facilitates easier analysis of the result plots. 

Since all games in the experiment were 4-player games and $\epsilon$
was set so that greedy selection will be used 25\% of the time, the
number of greedy move evaluations would be the same for both
MCE-$\epsilon$G/R and MCE-PG/R and should result in approximately the
same playout frequency for the two simulation strategies. This will
tell us how important accurate opponent modelling is in Kingdomino.

\begin{figure}[!t]
\centering
\includegraphics{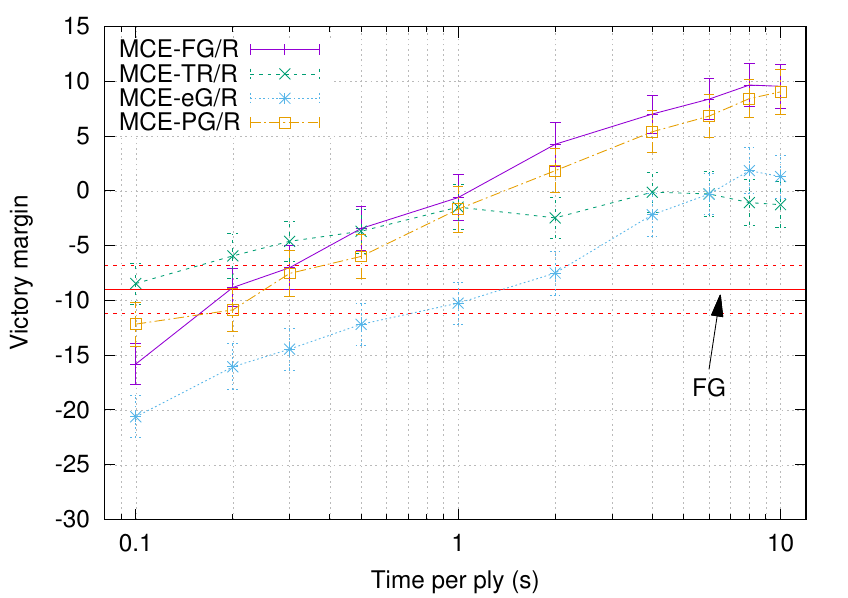}
\caption{Average victory margins against three FG opponents.}
\label{fig:experiment4_score_diffs}
\end{figure}

\begin{figure}[!t]
\centering
\includegraphics{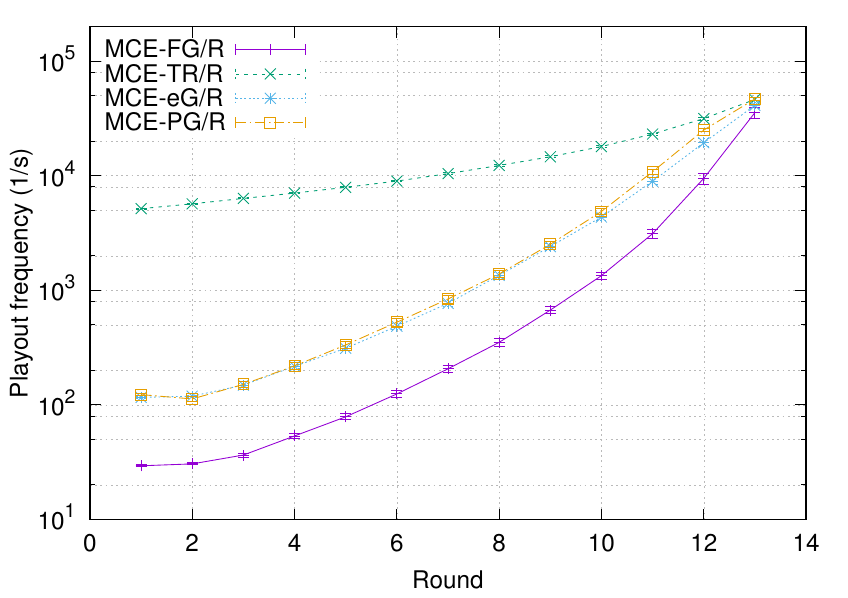}
\caption{Average playout frequency (200 games).}
\label{fig:experiment4_playouts}
\end{figure}

\figurename~\ref{fig:experiment4_score_diffs} shows the victory margin
under various time constraints for the different strategies (each
point represents 200 games). In addition to the Monte Carlo Evaluation
game strategies, the result from playing 200 games with an FG player
against three FG opponents is also shown (the solid red line with the
95\% confidence interval as dotted red
lines). \figurename~\ref{fig:experiment4_playouts} shows the number of
playouts per second for each playout policy.

The experiment shows that the FG evaluator is competitive to the
statistical evaluators under tight time constraints. It is comparable
to MCE-TR/R, and outperforms all the others, when the time is capped
to $0.1s$ per move. It also shows that the best knowledge-based
statistical evaluators need approximately $0.5-1s$ time per move for
the extra heuristic computations to pay off compared to selecting
playout moves randomly, but they consistently outperform the random
playout policy for move times $>1s$. It also shows that it is
more important to model the player's own move realistically than the
moves of the opponent. This is clear from the difference in performance
between MCE-PG/R and MCE-$\epsilon$G/R when having approximately the
same playout frequencies. Furthermore, if we compare MCE-PG/R to
MCE-FG/R we see that realistic opponent modelling is disadvantageous
for short ply times ($<0.2s$). This is natural since realistic
opponent modelling is costly and MCE-FG/R will only have time for few
playouts before selecting its move, while MCE-PG/R can produce more
playouts and have a better statistical sample when choosing its
move. However, once the number of playouts go up ($>0.1s$) we see that
realistic opponent modelling consistently outperforms the
player-greedy strategy, although not by much.

\subsection{Tree Search}

\begin{figure}[!t]
\centering
\includegraphics{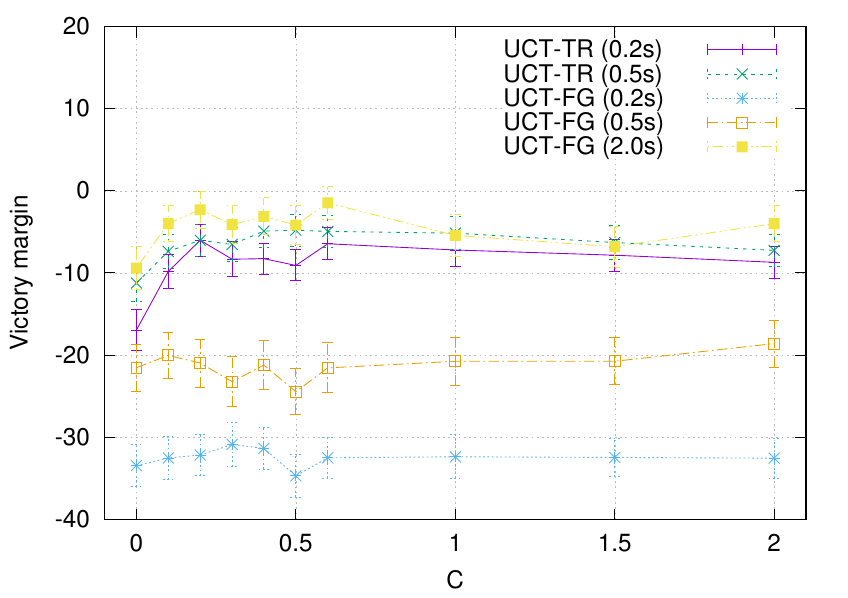}
\caption{Average victory margins against three FG opponents.}
\label{fig:experiment6_ucb_constant}
\end{figure}

We examined the UCB exploration constant $C$ by playing an UCT-TR and
an UCT-FG player against three FG players for various values of
$C$. The result is shown in
\figurename~\ref{fig:experiment6_ucb_constant}. The experiment shows
that $C = 0.6$ is a suitable value for players with many playouts per
ply and $C \ge 1.0$ for strategies with few playouts per ply. A theory
is that due to Kingdomino's frequent stochastic events, a move
requires numerous playouts to accumulate a representative reward. So
there is a risk of focusing the tree expansion on high-reward moves
before all moves get representative rewards. Therefore, players with
few playouts per ply should perform better with a higher exploration
constant.

\begin{figure}[!t]
\centering
\includegraphics{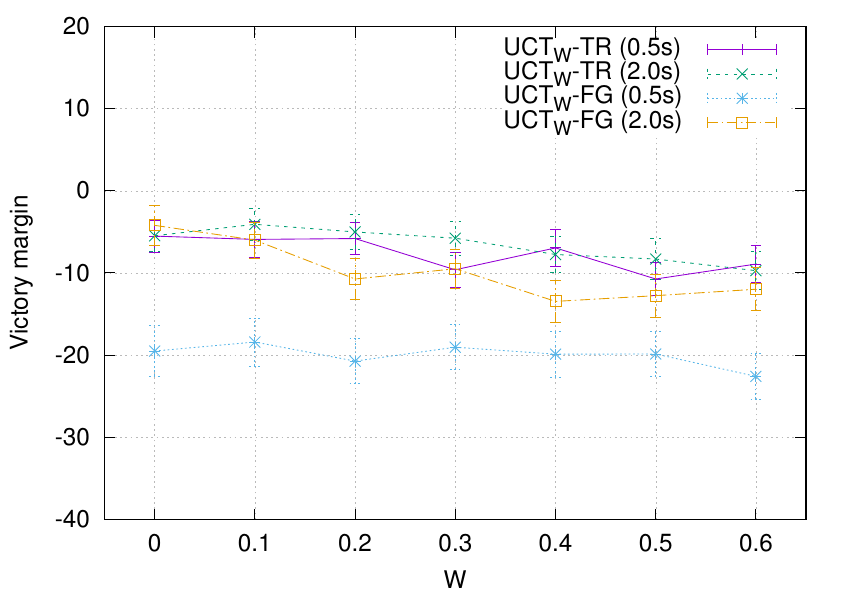}
\caption{Average victory margins against three FG opponents.}
\label{fig:experiment7_bias_constant}
\end{figure}

We also examined the impact constant $W$ for
progressive bias and progressive win bias by playing a
UCT\sub{W}-TR player and a UCT\sub{W}-FG player, both with $C=0.6$,
against three FG opponents for various values of $W$. The result is
shown in \figurename~\ref{fig:experiment7_bias_constant}. It shows
that we get the highest performance impact for $W=0.1\sim0.2$ and after
that the performance decreases with $W$.

\begin{table*}[!ht]
\begin{center}
\caption{Average victory margins for $200$ games against three FG opponents.}
\label{tab:experiment999}
\footnotesize{
\begin{tabular}{ll*{10}{r}}
\toprule
Strategy  & $\,$ & \multicolumn{10}{c}{Time per ply}   \\
\cmidrule{3-12}
       &  & $0.1$s  & $0.2$s & $0.3$s & $0.5$s & $1.0$s & $2.0$s & $4.0$s & $6.0$s & $8.0$s & $10.0$s  \\
\midrule
FG                      & &  \five{-9.0} &  \ten{-9.0} &  \ten{-9.0} &   \ten{-9.0} &        -9.0  &       -9.0  &       -9.0  &       -9.0  &       -9.0  &       -9.0  \\
MCE-TR/R                & &  \best{-8.5} & \best{-5.9} & \best{-4.6} &  \five{-3.7} &  \five{-1.5} &       -2.5  &       -0.1  &       -0.3  &       -1.1  &       -1.3  \\
MCE-FG/R                & &       -15.8  &  \ten{-8.8} & \five{-7.0} &  \best{-3.4} &  \best{-0.6} &  \best{4.3} &  \best{7.0} &  \best{8.4} &  \best{9.7} &  \best{9.6} \\
MCE-PG/R                & &  \ten{-12.2} & \ten{-10.9} &  \ten{-7.5} &  \five{-6.0} &  \five{-1.7} &  \five{1.9} &  \five{5.4} &  \five{6.8} &  \five{8.4} &  \five{9.0} \\
MCE-$\epsilon$G/R       & &       -20.6  &      -16.0  &      -14.5  &       -12.2  &       -10.3  &       -7.5  &       -2.2  &       -0.3  &        1.9  &        1.3  \\
UCT-TR                  & &  \ten{-13.5} & \five{-6.4} & \five{-7.3} &  \five{-4.9} &   \ten{-5.5} &       -4.4  &       -3.5  &       -5.2  &       -7.2  &       -4.4  \\
UCT-FG                  & &       -38.3  &      -32.5  &      -29.4  &       -21.6  &       -12.0  &  \ten{-1.5} &       -0.2  &   \ten{3.5} &   \ten{4.0} &   \ten{3.9} \\
UCT-PG                  & &       -25.8  &      -20.7  &      -15.5  &       -15.3  &       -13.9  &      -10.4  &       -7.1  &       -7.4  &       -6.2  &       -4.1  \\
UCT-$\epsilon$G         & &       -33.3  &      -24.0  &      -16.2  &       -15.7  &        -9.0  &       -7.0  &       -3.1  &       -4.0  &       -2.6  &       -1.3  \\
UCT\sub{B}-TR           & & \five{-10.1} & \five{-7.4} & \five{-6.1} &   \ten{-7.9} &   \ten{-4.6} &       -6.7  &       -4.8  &       -4.3  &       -2.9  &       -4.5  \\
UCT\sub{B}-FG           & &       -39.8  &      -30.6  &      -29.7  &       -21.6  &       -11.7  &       -5.9  &       -0.1  &   \ten{3.2} &   \ten{3.2} &        1.4  \\
UCT\sub{B}-PG           & &       -25.5  &      -20.8  &      -19.7  &       -15.4  &       -13.4  &       -9.1  &       -7.6  &       -6.3  &       -4.5  &      -12.4  \\
UCT\sub{B}-$\epsilon$G  & &       -31.5  &      -25.2  &      -20.2  &       -16.3  &       -10.7  &       -8.1  &       -4.1  &       -2.6  &       -1.9  &       -2.9  \\
UCT\sub{W}-TR           & &  \ten{-11.4} & \five{-6.7} & \five{-7.3} &  \five{-5.9} &   \ten{-4.6} &       -4.1  &       -5.8  &       -5.0  &       -4.0  &       -4.5  \\
UCT\sub{W}-FG           & &       -42.6  &      -33.0  &      -30.3  &       -18.4  &       -13.9  &       -6.0  &       -2.5  &        0.6  &        1.2  &        1.4  \\ 
UCT\sub{W}-PG           & &       -29.2  &      -24.3  &      -20.0  &       -19.7  &       -15.4  &      -16.9  &      -13.1  &      -12.2  &      -13.9  &      -12.4  \\
UCT\sub{W}-$\epsilon$G  & &       -30.5  &      -23.0  &      -22.8  &       -16.6  &       -13.2  &       -6.6  &       -5.4  &       -3.1  &       -2.7  &       -2.9  \\
\bottomrule
\end{tabular}
}
\end{center}
\end{table*}

\subsection{Comparing Strategies}

Table~\ref{tab:experiment999} shows the performance of all strategies
for $200$ games played against three FG opponents. The $95$\%
confidence intervals are in the range $[3.5, 6.0]$ for all entries,
with the majority near the lower limit. The highest performer for each
time constraint is marked by a dark blue box. Performances within
$5\%$ ($10\%$) of the best are marked by a light (lighter) blue
box. The UCB exploration constant was set to $C=0.6$ for all UCT
strategies and the the bias impact factor was set to $W=0.1$ for
UCT\sub{B}-* and UCT\sub{W}-*.

The results show that for each time constraint the best MCE variant
consistently outperforms all variants of UCT. A possible theory is
that UCT is hampered by its WDL scoring function, but further
experiments verifying this hypothesis is outside the scope of this
paper. The true random playout policy variant (MCE-TR/R) excels for
short ply times $t<0.5s$. After that the full greedy playout policy
variant (MCE-FG/R) gets enough time each ply to produce rewards
representative enough to reliably select trajectories in the game tree
that outperform the the random playout policy, in spite of the
significantly higher playout frequency of the random playout
policy. The MCE-PG performs almost on par with MCE-FG which indicates
that allocating time for accurate opponent modelling only has a small
gain compared to using random move selection for the opponents.

The results also show that the UCT enhancements improve the results for
tight time constraints ($t<0.2s$), which is expected due to few
playouts, but are otherwise on par with regular UCT.

\section{Conclusions and Future Work}
\label{sec:conclusions}

This paper introduces Kingdomino as an interesting game to study for
game playing. The shallow game tree and relatively limited interaction
between players of Kingdomino combined with the stochastic nature and
possibility to evaluate partial game states is particularly
interesting. The results indicate that for games such as Kingdomino,
MCE is superior to UCT, which would infer new recommendations on the
suitability of MCE for games of similar complexity. This is especially
interesting, given that an MCE evaluator is significantly easier to
implement correctly and efficiently than full UCT.

The \emph{player-greedy} playout policy is surprisingly effective,
balancing exploration power with (expensive) local evaluation. Our
belief is that this is due to the limited (but not insignificant)
interaction among players in Kingdomino, but further experiments in
other games are needed to verify this hypothesis. The
\emph{progressive win bias} selection improvement shows promise as a
way to combine a heuristic evaluation with the current knowledge
gained from the exploration, but further experiments in other settings
better suited for the UCT is needed to analyse its impact.

Our evaluation uses thorough systematic examination of all constants
involved to avoid the presence of magic numbers which frequently
occur without explanation in many similar papers in the
field. It also uses new and illuminating graphs for showing the impact
of different choices. In particular, the usage of victory margin in
favour of win percentages is very powerful for a multi player score
maximization game such as Kingdomino. These graphs have helped us gain
new insights into both the game and how our strategies perform.

For future work one MCTS enhancement alternative could be a
learning heuristic that keep offline information on the success of
placement positions for different kingdom patterns. Experienced human
players tend to place dominos in a structured pattern to avoid single
tile holes in the kingdom. It would also be interesting to implement
agents using completely different strategies such as deep
reinforcement learning.

The code for the Kingdomino game server can be downloaded from
\texttt{https://github.com/mratin/kdom-ai}, and the AI implementations
can be downloaded from \texttt{https://github.com/mgedda/kdom-ai}.


\section*{Acknowledgements}

We thank all participants at Tomologic who implemented agents and
discussed strategies with us.

\bibliographystyle{splncs}
\bibliography{references}

\end{document}